%% file: main.tex
\documentclass{article}
\usepackage{spconf,amsmath,graphicx}
\usepackage{amsmath}  
\usepackage{amsfonts} 

\usepackage{multirow}
\usepackage{booktabs}
\usepackage{xcolor}
\definecolor{cvprblue}{rgb}{0.21,0.49,0.74}
\usepackage[pagebackref,breaklinks,colorlinks,citecolor=cvprblue]{hyperref}

\usepackage{enumitem}
\setlist{nosep, leftmargin=14pt}

\usepackage{mwe} 

\title{Optimised ProPainter for Video Diminished Reality Inpainting}
\name{Pengze Li $^{*}$  \qquad  Lihao Liu $^{*}$  \qquad Carola-Bibiane Schönlieb \qquad  Angelica I Aviles-Rivero }
\address{Univerisity of Cambridge}

\begin{document}
%
\maketitle

\input{section/abstract}

\input{section/introduction}

\input{section/method}

\input{section/experiments}

\input{section/conclusion}

\bibliographystyle{IEEEbib}
\bibliography{strings,refs}

\end{document}

%% file: section/abstract.tex
\begin{abstract}
    In this paper, part of the DREAMING Challenge - Diminished Reality for Emerging Applications in Medicine through Inpainting, we introduce a refined video inpainting technique optimized from the ProPainter method to meet the specialised demands of medical imaging, specifically in the context of oral and maxillofacial surgery. Our enhanced algorithm employs the zero-shot ProPainter, featuring optimized parameters and pre-processing, to adeptly manage the complex task of inpainting surgical video sequences, without requiring any training process. It aims to produce temporally coherent and detail-rich reconstructions of occluded regions, facilitating clearer views of operative fields. The efficacy of our approach is evaluated using comprehensive metrics, positioning it as a significant advancement in the application of diminished reality for medical purposes. 
\end{abstract}

\renewcommand{\thefootnote}{\fnsymbol{footnote}}
\footnotetext[1]{\,\,The two authors contributed equally to this work.}

%% file: section/introduction.tex
\section{Introduction}

Image inpainting~\cite{bertalmio2000image}, a fundamental technique in computer vision, restores missing or damaged regions in images by generating a complete output. This process is crucial not only in still photography but extends significantly into the video setting. Video inpainting e.g.,~\cite{zhou2023propainter} is more challenging; as it requires maintaining temporal consistency across frames to avoid visible discrepancies. The advent of deep learning-based inpainting~\cite{quan2024deep} methods has revolutionised this field, enabling more sophisticated applications such as art restoration, photo editing, and now, video editing for scenes where continuity and consistency are essential. These advancements have made video inpainting an interesting tool in film production, video restoration, and augmented reality applications, where integration of real and virtual elements is paramount. By addressing both spatial and temporal gaps, video inpainting ensures a cohesive visual narrative.

Video inpainting has been widely explored in the literature. A variety of techniques have focused on patch-based variational approaches e.g.,~\cite{wexler2007space, liu2022pc}. While they have achieved good results, the major constraints are the large computational cost of the optimisation scheme and the limited consistency gain. Deep learning-based techniques have emerged as a means to mitigate such disadvantages, which can be broadly divided into 3D CNN techniques, shift methods, flow-guided methods, and attention techniques.
A set of techniques has focused on improving 3D CNN techniques~\cite{wang2019video,chang2019free,hu2020proposal,liu2019probabilistic,liu2020psi}, typically employing a two-stage solution. While shift methods~\cite{ouyang2021internal,ke2021occlusion} appear to mitigate the computational cost of 3D convolution operations, flow-based techniques have demonstrated outstanding temporal consistency; however, the results greatly depend on an accurate approximation of optical flow~\cite{lao2021flow, kang2022error}. More recently, attention-based methods~\cite{kim2019deep, li2023local,liu2023scotch} have led in performance by integrating both short and long-term video information.

In this work, we specifically explore the application of video inpainting within a medical context, focusing on oral and maxillofacial surgery~\cite{challengepaper1,luijten20233d}. We utilise simulated surgical video sequences from the perspective of an operating surgeon. These sequences provide a detailed view of the surgical environment, including elements such as the surgeon's hands and surgical tools. Such elements may occasionally obscure critical areas of the patient's face or other important parts of the scene, necessitating their algorithmic removal to restore an unobstructed view of the surgical site. Hence, \textit{our primary objective is to use video inpainting techniques to regenerate the underlying background in surgical imagery.} To achieve this, we proposed a zero-shot ProPainter model, optimized with refined parameters and preprocessing techniques. Experimental results indicate that our model ranks 1$^{st}$ in Phase 1 of the DREAMING challenge.

%% file: section/method.tex
\section{Methodology}

In this section, we first introduce the core components of the ProPainter model. Next, we  discuss on how we optimise the model in a zero-shot manner, incorporating pre-processing techniques and describing the experimental setup.

\begin{figure*}[htb]
  \centering
  \includegraphics[width=0.9\textwidth]{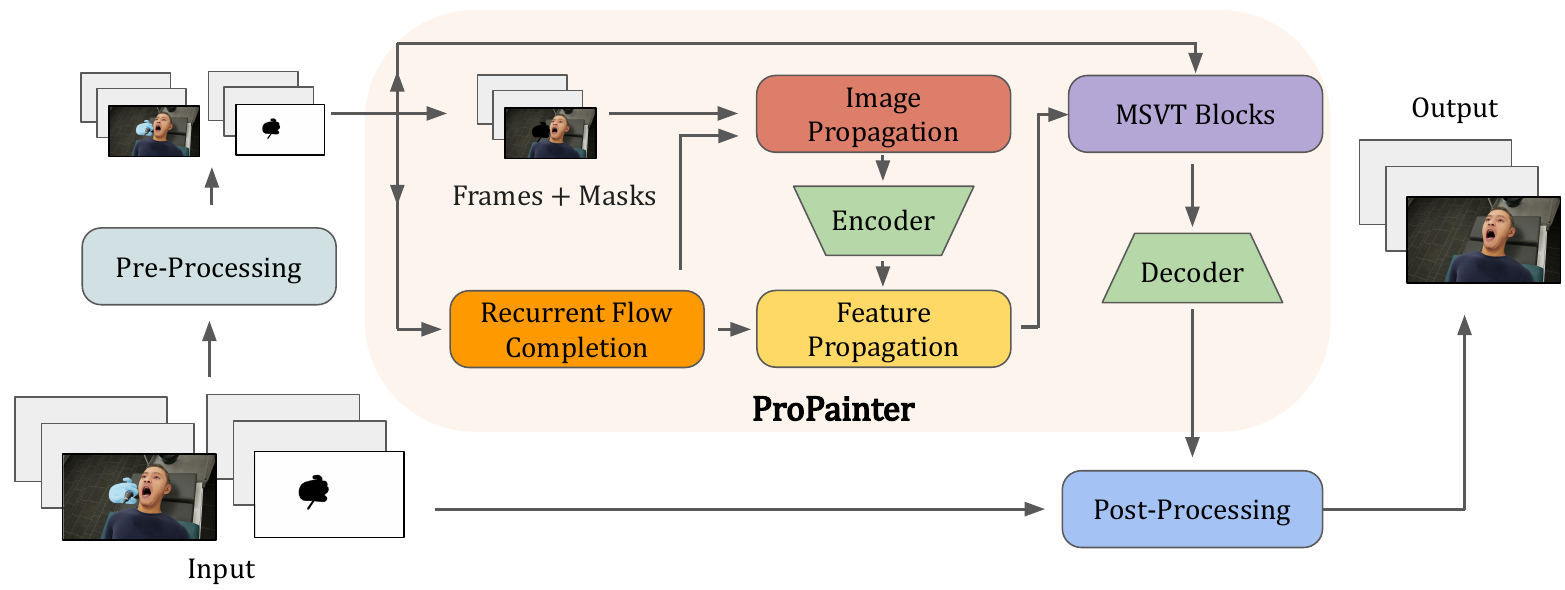}
  \caption{Schematic of the ProPainter inpainting pipeline in our framework: This diagram illustrates the workflow of our ProPainter-based video inpainting. MSVT: Mask-guided Sparse Video Transformer.}
  \label{fig:yourlabel}
\end{figure*}

\subsection{ProPainter}

ProPainter \cite{zhou2023propainter} combines enhanced propagation mechanisms with transformer, ideal for addressing the complexities of maintaining temporal and textural consistency in video sequences.
It operates by first employing its propagation modules to estimate motion and texture information from adjacent frames into the regions requiring inpainting. Then, the transformer module of the network then processes these propagated estimates to refine the inpainting output. 

\textbf{Recurrent Flow Completion.}
In video inpainting, pre-trained flow completion modules play a critical role by simplifying the task of direct RGB content filling and enhancing temporal coherence~\cite{zhang2022inertia}.
It encode motion flows \( F_t \) into downsampled features \( f_t \) with an 8x downsampling ratio, using deformable convolution (DCN) \cite{zhu2019deformable} for bidirectional flow propagation. For backward flow propagation, the network processes the concatenated features \( c(f_t, \hat{f}_{t+1}) \)—with \( \hat{f}_{t+1} \) representing the subsequent frame's features—through convolutions that calculate DCN offsets \( o_{t \rightarrow t+1} \) and modulation masks \( m_{t \rightarrow t+1} \). Following the ProPainter~\cite{zhou2023propainter}, this process then reads:
\begin{equation} \label{eq:1}
\hat{f_t} = \mathcal{R} \big (\mathcal{D} (\hat{f}_{t+1}; o_{t\rightarrow t+1}, m_{t\rightarrow t+1}), f_t\big )
\end{equation}
where \( \mathcal{D}( \cdot ) \) represents the deformable convolution operation that adapts the position and scale of the convolution kernels based on learned offsets and modulation parameters, facilitating dynamic feature alignment. \( \mathcal{R}( \cdot ) \) denotes the convolutional layers that fuse the dynamically aligned features from \( \mathcal{D} \) with the current frame features, enhancing the final quality and coherence of the reconstructed flow information. A decoder then reconstructs the completed flows \( \hat{F}_t \).

\textbf{Image Propagation and Feature Propagation.}
Dual-domain propagation in video inpainting is critical for maintaining spatial and temporal consistency, which contains image propagation and feature propagation. The image propagation uses a warping operation, combined with the calculated reliable areas \(A_r\), to update the current frame.
\begin{equation}
    \hat{X}_{t} = \mathcal{W}(X_{t+1}, \hat{F}_{t \rightarrow t+1}) * A_r + X_t * (1 - A_r)
\end{equation}
where \( \mathcal{W}( \cdot ) \) denotes the warping operation. 

Feature propagation is achieved through a flow-guided deformable alignment mechanism. An image encoder extracts features from the video sequence, which are then aligned using flow information.  Following ProPainter~\cite{zhou2023propainter}, we have:
\begin{equation} 
    \hat{e}_t = \mathcal{R} \big(\mathcal{D}(\hat{e}_{t+1}; \hat{F}_{t \rightarrow t+1}^{\downarrow} + \widetilde{o}_{t \rightarrow t+1}, m_{t \rightarrow t+1}), f_t\big)
\end{equation}
where \( \mathcal{D}( \cdot ) \) represents deformable convolution and \( \mathcal{R}( \cdot ) \) is the same as define in Equation~\ref{eq:1}. This method leverages completed flows and additional mask conditions to improve alignment accuracy and propagation reliability, focusing on areas where previous propagation may be less reliable. This dual-domain approach ensures that both global image and local feature information are propagated efficiently, enhancing the quality and coherence of the inpainted video sequences.

\textbf{Mask-guided Sparse Video Transformer (MSVT) Blocks.}
Video Transformers have shown excellent performance in video inpainting but are often limited by their high computational and memory demands. To address these limitations, ProPainter~\cite{zhou2023propainter} introduce a novel sparse video Transformer that extends the window-based approach, enhancing efficiency without compromising effectiveness. A strategy where attention is selectively applied only to query windows that intersect with masked regions.

\subsection{Optimising ProPainter}
As a transformer-based method, ProPainter inevitably requires significant memory consumption and time to execute. This initially did not align with the real-time diminished reality solution desired by the organizers. To address this, we have made several enhancements to reduce both the memory and time requirements.

\textbf{Pre-Processing.} Before processing with ProPainter, we reduce the resolution of the overall video sequence and mask sequence. This step directly decreases the computational load while maintaining sufficient detail for effective inpainting. We also slightly expand the mask to encompass more information, enhancing the quality of the inpainting by providing more context to the algorithm.

\textbf{Memory Management.} We optimised the memory management process to offload unnecessary tensors to the CPU rather than storing them on the GPU. This adjustment ensures that only the video clips currently being processed are held on the GPU, and intermediate variables generated during the process are promptly deleted.

\textbf{Post-Processing.} After inpainting, the filled image is resized back to its original dimensions. We then seamlessly integrate this image back into the original video sequence, replacing the masked areas.

\textbf{Experimental Setup.} The ProPainter model was initialised with its pre-trained state, taking advantage of its comprehensive training on a wide array of video data, including YouTube-VOS \cite{xu2018youtube} and DAVIS \cite{perazzi2016benchmark}, to ensure robust generalization to the diverse scenarios encountered in the challenge. In our submission, rather than fine-tuning our model on the training dataset, we adopt a zero-shot learning approach using pre-trained weights. For preprocessing, we reduce the resolution by 30\%, which enables our ProPainter model to function effectively within the memory constraints of a T4 GPU, equipped with 16GB of RAM. This adjustment facilitates the processing of nearly 1,000 video frames within the prescribed 20-minute limit. During the inpainting phase, we carefully select 18 neighboring frames to provide temporal context, and designate global reference frames at every 20-frame interval to ensure coherence and quality in the output.

%% file: section/experiments.tex
\section{Experimental Results}

\subsection{Dataset and Evaluation Metrics}

The dataset employed in this study was provided by the organizers of the associated challenge, consisting of 100 diverse scenes from oral and maxillofacial surgeries, depicted in approximately 1,000 images each at a resolution of 1280 $\times$ 720 pixels across three color channels. Images include both obstructed and unobstructed views, with the former marked by binary masks. The testing of the dataset will take place on 2 unseen scenes during the Preliminary Container Testing Phase and an additional 10 scenes in the Final Test Phase. 

The model's performance is assessed using four metrics: Learned Perceptual Image Patch Similarity (LPIPS), Frechet Inception Distance (FID) for feature-wise evaluation, and Mean Absolute Error (MAE) and Peak Signal to Noise Ratio (PSNR) for pixel-wise evaluation. These metrics will determine the final rankings in the challenge. Specifically, weighted contributions from LPIPS and FID are used to calculate the Consistency Error (C-Error), while weighted MAE and PSNR are utilized to determine the Accuracy Error (A-Error). Given the inherent challenges of inpainting as an ill-posed problem with multiple viable visual outcomes, the challenge prioritizes consistency and feature-based evaluations. Rankings prioritize the lowest values of C-Error, followed by A-Error, to emphasize their significance.

\subsection{Numerical Results}

\input{table/table_1}

In our evaluation, we benchmark our approach against the latest state-of-the-art (SOTA) generative model, Stable Diffusion~\cite{rombach2022high}. We conduct this comparison on the challenge training dataset using cross-validation.

Table~\ref{table_1_local_results} provides a summary of the evaluation metrics, including Weighted-FID (W-FID), Weighted Mean Absolute Error (W-MAE), Weighted Peak Signal-to-Noise Ratio (W-PSNR), Weighted Learned Perceptual Image Patch Similarity (W-LPIPS), Accuracy Error (A-Error), and Consistency Error (C-Error) for our proposed technique compared to the Stable Diffusion methods. Our method outperforms across all six metrics, indicating a robust improvement over Stable Diffusion methods. We attribute this enhancement to our video-level methodology, ProPainter, which demonstrates superior performance, particularly in the context of dynamic scenes, as opposed to the image-level focus of Stable Diffusion. This suggests that our video-oriented approach is more effective in capturing and generating temporally consistent and perceptually accurate results.

\begin{figure}[t!]
  \centering
  \includegraphics[width=\linewidth, keepaspectratio]{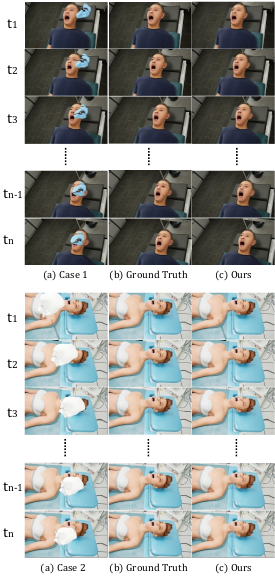}
  \caption{Visual comparison of video inpainting on surgical sequences: selected frames demonstrate our method's effectiveness across different time points $(t1, t2, t3, t_{n-1}, t_n)$. 'Case 1' and 'Case 2' compare the original frames with obstructions ('Ground Truth') against the inpainted results ('Ours').}
  \label{fig:out}\vspace{-0.2cm}
\end{figure}

\textbf{Leaderboard on Grand Challenge.} Table \ref{table_2_online_results} showcases the leaderboard results from the Phase 1 - Preliminary Container Testing Phase of the challenge dataset. Our encapsulated algorithm was submitted for blind evaluation, yielding the results presented. At the time of this paper's submission, our ProPainter-based approach leads the competition, achieving the 1$^{st}$ rank. This reflects the robustness of our algorithm in the context of the challenge's specific tasks and constraints.

\input{table/table_2}

\subsection{Visualisation Results}

In Figure~\ref{fig:out} we presents a set of visualisations that compare the input videos, ground truth, and our inpainted predictions. Figure labels (a), (b), and (c) correspond to the input videos, ground truth, and inpainted predictions, respectively.

In our input videos (column a), the face and body of the patient are intermittently obscured by simulated hands and medical instruments, designed to mimic realistic surgical scenarios. Columns (b) and (c) display the ground truth and our algorithm's inpainted predictions, illustrating the effectiveness of our method in reconstructing the structure of the patient's face and body obstructed by the hands.

Two scenes with representative frames were specifically selected where hands obstruct the face, highlighting the challenges and performance of our reconstruction technique. In the first scene, indexed at \( n = 861 \), and in the second, at \( n = 506 \), we exhibit the visualization results at every third frame, capturing the dynamics as obstructions move across the scene. This approach demonstrates the temporal consistency and robustness of our model in handling moving obstructions within the video frames.

%% file: table/table_1.tex
\begin{table}[t!]
\centering
\resizebox{\linewidth}{!}{
    \begin{tabular}{c|cccccc} \hline \toprule
        \multicolumn{1}{c|}{\multirow{2}{*}{\textsc{}}}  & \multicolumn{6}{c}{\textsc{Evaluation Metrics}}                                                         \\ \cline{2-7}
        & $\text{W-FID}$      & $\text{W-MAE}$  & $\text{W-PSNR}$   & $\text{W-LPIPS}$   & $\text{A-Error}$  & $\text{C-Error}$                                        \\ \midrule
       
        \multicolumn{1}{l|}{SD\,\,\,\,\,\,\,\,\,\,\,\,}                        & 0.314    & 0.297    & 0.278     & 0.351     & 0.287     & 0.333                                         \\ \midrule
        \multicolumn{1}{l|}{Ours}                                              & \textbf{0.224} & \textbf{0.293} & \textbf{0.232} & \textbf{0.329} & \textbf{0.263} & \textbf{0.276}     \\ \bottomrule
        \end{tabular}
}
\caption{Comparative analysis of our ProPainter-based algorithm and Stable Diffusion using cross-validation on the local challenge test dataset. The evaluation metrics, where lower scores indicate better performance, include: Weighted-FID (W-FID), Weighted Mean Absolute Error (W-MAE), Weighted Peak Signal-to-Noise Ratio (W-PSNR), Weighted Learned Perceptual Image Patch Similarity (W-LPIPS), Accuracy Error (A-Error), and Consistency Error (C-Error).
}
\label{table_1_local_results}
\end{table}

%% file: table/table_2.tex
\begin{table}[t!]
\centering
\resizebox{\linewidth}{!}{
    \begin{tabular}{c|cccccc} \hline \toprule
        \multicolumn{1}{c|}{\multirow{2}{*}{\textsc{}}}                           & \multicolumn{6}{c}{\textsc{Evaluation Metrics}}                                                         \\ \cline{2-7}
                                                                                & $\text{W-FID}$      & $\text{W-MAE}$  & $\text{W-PSNR}$   & $\text{W-LPIPS}$   & $\text{A-Error}$  & $\text{C-Error}$                                            \\ \midrule
        \multicolumn{1}{l|}{Baseline}                                           & 0.792    & \textbf{0.257}    & 0.255     & 0.791     & 0.256     & 0.792                                         \\ \midrule
        \multicolumn{1}{l|}{Team 1}                        & 0.075    & 0.260    & 0.235     & 0.349     & 0.247     & 0.212                                         \\ 
        \multicolumn{1}{l|}{Team 2}                      & 0.208    & 0.263    & 0.244     & 0.439     & 0.253     & 0.324\\ 
        \multicolumn{1}{l|}{Team 3}                      & 0.079    & 0.259    & \textbf{0.218}     & 0.292     & \textbf{0.239}     & 0.186                                         \\ \midrule
        \multicolumn{1}{l|}{Ours}                                              & \textbf{0.071} & 0.259 & 0.221 & \textbf{0.287} & 0.240 & \textbf{0.179}     \\ \bottomrule
        \end{tabular}
}
\caption{Comparative Analysis of Our ProPainter-Based Algorithm and Algorthms from Other Team on the Phase 1 - Preliminary Container Testing Phase. 
}
\label{table_2_online_results}
\end{table}

%% file: section/conclusion.tex
\section{Conclusion}
We adapted the ProPainter framework for the DREAMING Challenge. It has led to  advancements in diminished reality for medical video inpainting. Our tailored approach has effectively addressed the unique requirements of surgical imagery, enabling the clear reconstruction of occluded areas while maintaining temporal coherence. The successful application in oral and maxillofacial surgeries exemplifies the potential of our technique to enhance visual clarity for medical practitioners.

%% file: main.bbl
\begin{thebibliography}{10}

\bibitem{bertalmio2000image}
Marcelo Bertalmio, Guillermo Sapiro, Vincent Caselles, and Coloma Ballester,
\newblock ``Image inpainting,''
\newblock in {\em Proceedings of the 27th annual conference on Computer graphics and interactive techniques}, 2000, pp. 417--424.

\bibitem{zhou2023propainter}
Shangchen Zhou, Chongyi Li, Kelvin~CK Chan, and Chen~Change Loy,
\newblock ``Propainter: Improving propagation and transformer for video inpainting,''
\newblock in {\em International Conference on Computer Vision}, 2023, pp. 10477--10486.

\bibitem{quan2024deep}
Weize Quan, Jiaxi Chen, Yanli Liu, Dong-Ming Yan, and Peter Wonka,
\newblock ``Deep learning-based image and video inpainting: A survey,''
\newblock {\em International Journal of Computer Vision}, pp. 1--34, 2024.

\bibitem{wexler2007space}
Yonatan Wexler, Eli Shechtman, and Michal Irani,
\newblock ``Space-time completion of video,''
\newblock {\em IEEE Transactions on pattern analysis and machine intelligence}, vol. 29, no. 3, pp. 463--476, 2007.

\bibitem{liu2022pc}
Lihao Liu, Zhening Huang, Pietro Li{\`o}, Carola-Bibiane Sch{\"o}nlieb, and Angelica~I Aviles-Rivero,
\newblock ``Pc-swinmorph: Patch representation for unsupervised medical image registration and segmentation,''
\newblock {\em arXiv preprint arXiv:2203.05684}, 2022.

\bibitem{wang2019video}
Chuan Wang, Haibin Huang, Xiaoguang Han, and Jue Wang,
\newblock ``Video inpainting by jointly learning temporal structure and spatial details,''
\newblock in {\em AAAI}, 2019, pp. 5232--5239.

\bibitem{chang2019free}
Ya-Liang Chang, Zhe~Yu Liu, Kuan-Ying Lee, and Winston Hsu,
\newblock ``Free-form video inpainting with 3d gated convolution and temporal patchgan,''
\newblock in {\em International Conference on Computer Vision}, 2019.

\bibitem{hu2020proposal}
Yuan-Ting Hu, Heng Wang, Nicolas Ballas, Kristen Grauman, and Alexander~G Schwing,
\newblock ``Proposal-based video completion,''
\newblock in {\em European Conference of computer vision}, 2020, pp. 38--54.

\bibitem{liu2019probabilistic}
Lihao Liu, Xiaowei Hu, Lei Zhu, and Pheng-Ann Heng,
\newblock ``Probabilistic multilayer regularization network for unsupervised 3d brain image registration,''
\newblock in {\em Medical Image Computing and Computer Assisted Intervention}. Springer, 2019, pp. 346--354.

\bibitem{liu2020psi}
Lihao Liu, Xiaowei Hu, Lei Zhu, Chi-Wing Fu, Jing Qin, and Pheng-Ann Heng,
\newblock ``$\psi$-net: Stacking densely convolutional lstms for sub-cortical brain structure segmentation,''
\newblock {\em IEEE transactions on medical imaging}, vol. 39, no. 9, pp. 2806--2817, 2020.

\bibitem{ouyang2021internal}
Hao Ouyang, Tengfei Wang, and Qifeng Chen,
\newblock ``Internal video inpainting by implicit long-range propagation,''
\newblock in {\em Computer Vision and Pattern Recognition (CVPR)}, 2021, pp. 14579--14588.

\bibitem{ke2021occlusion}
Lei Ke, Yu-Wing Tai, and Chi-Keung Tang,
\newblock ``Occlusion-aware video object inpainting,''
\newblock in {\em Proceedings of the IEEE/CVF International Conference on Computer Vision}, 2021, pp. 14468--14478.

\bibitem{lao2021flow}
Dong Lao, Peihao Zhu, Peter Wonka, and Ganesh Sundaramoorthi,
\newblock ``Flow-guided video inpainting with scene templates,''
\newblock in {\em Computer Vision and Pattern Recognition (CVPR)}, 2021, pp. 14599--14608.

\bibitem{kang2022error}
Jaeyeon Kang, Seoung~Wug Oh, and Seon~Joo Kim,
\newblock ``Error compensation framework for flow-guided video inpainting,''
\newblock in {\em European conference on computer vision}. Springer, 2022, pp. 375--390.

\bibitem{kim2019deep}
Dahun Kim, Sanghyun Woo, Joon-Young Lee, and In~So Kweon,
\newblock ``Deep video inpainting,''
\newblock in {\em Computer Vision and Pattern Recognition}, 2019, pp. 5792--5801.

\bibitem{li2023local}
Chen Li, Li~Song, Rong Xie, and Wenjun Zhang,
\newblock ``Local bidirection recurrent network for efficient video deblurring with the fused temporal merge module,''
\newblock {\em ACM Transactions on Multimedia Computing, Communications and Applications}, pp. 1--18, 2023.

\bibitem{liu2023scotch}
Lihao Liu, Jean Prost, Lei Zhu, Nicolas Papadakis, Pietro Li{\`o}, Carola-Bibiane Sch{\"o}nlieb, and Angelica~I Aviles-Rivero,
\newblock ``Scotch and soda: A transformer video shadow detection framework,''
\newblock in {\em Computer Vision and Pattern Recognition (CVPR)}, 2023, pp. 10449--10458.

\bibitem{challengepaper1}
Christina Gsaxner, Shohei. Mori, Gijs. Luijten, Viet~Duc Vu, Timo van Meegdenburg, Gabriele~A. Krombach, Jens Kleesiek, Ulrich Eck, Nassir Navab, Yan Guo, Xiaojun Chen, Frank Hölzle, Behrus Puladi, and Jan Egger,
\newblock ``Diminished reality for emerging applications in medicine through inpainting,''
\newblock in {\em International Symposium on Biomedical Imaging}, 2024.

\bibitem{luijten20233d}
Gijs Luijten, Christina Gsaxner, Jianning Li, Antonio Pepe, Narmada Ambigapathy, Moon Kim, Xiaojun Chen, Jens Kleesiek, Frank H{\"o}lzle, Behrus Puladi, et~al.,
\newblock ``3d surgical instrument collection for computer vision and extended reality,''
\newblock {\em Scientific Data}, vol. 10, no. 1, pp. 796, 2023.

\bibitem{zhang2022inertia}
Kaidong Zhang, Jingjing Fu, and Dong Liu,
\newblock ``Inertia-guided flow completion and style fusion for video inpainting,''
\newblock in {\em Computer Vision and Pattern Recognition (CVPR)}, 2022, pp. 5982--5991.

\bibitem{zhu2019deformable}
Xizhou Zhu, Han Hu, Stephen Lin, and Jifeng Dai,
\newblock ``Deformable convnets v2: More deformable, better results,''
\newblock in {\em Computer Vision and Pattern Recognition (CVPR)}, 2019, pp. 9308--9316.

\bibitem{xu2018youtube}
Ning Xu, Linjie Yang, Yuchen Fan, Dingcheng Yue, Yuchen Liang, Jianchao Yang, and Thomas Huang,
\newblock ``Youtube-vos: A large-scale video object segmentation benchmark,''
\newblock {\em arXiv preprint arXiv:1809.03327}, 2018.

\bibitem{perazzi2016benchmark}
Federico Perazzi, Jordi Pont-Tuset, Brian McWilliams, Luc Van~Gool, Markus Gross, and Alexander Sorkine-Hornung,
\newblock ``A benchmark dataset and evaluation methodology for video object segmentation,''
\newblock in {\em Computer Vision and Pattern Recognition (CVPR)}, 2016, pp. 724--732.

\bibitem{rombach2022high}
Robin Rombach, Andreas Blattmann, Dominik Lorenz, Patrick Esser, and Bj{\"o}rn Ommer,
\newblock ``High-resolution image synthesis with latent diffusion models,''
\newblock in {\em Computer Vision and Pattern Recognition (CVPR)}, 2022, pp. 10684--10695.

\end{thebibliography}
